# Knowledge Representation for Robots through Human-Robot Interaction

E. Bastianelli, D. Bloisi, R. Capobianco, G. Gemignani, L. Iocchi, D. Nardi

Dept. Computer, Control and Management Engineering
Sapienza University of Rome, Italy
`<lastname>@dis.uniroma1.it`

**Abstract.** The representation of the knowledge needed by a robot to perform complex tasks is restricted by the limitations of perception. One possible way of overcoming this situation and designing "knowledgeable" robots is to rely on the interaction with the user. We propose a multi-modal interaction framework that allows to effectively acquire knowledge about the environment where the robot operates. In particular, in this paper we present a rich representation framework that can be automatically built from the metric map annotated with the indications provided by the user. Such a representation, allows then the robot to ground complex referential expressions for motion commands and to devise topological navigation plans to achieve the target locations.

**Keywords:** Knowledge Representation, Robotics, Human-Robot Interaction

## 1 Introduction

Robots are expected to become consumer products. However, there is still a gap in terms of user expectations and robot functionality. A key limiting factor is the lack of knowledge and awareness of the robot on the problem to be solved and on the operational environment. The difficulties arise from several sources: the current capabilities of perception systems, the difficulty of communicating with humans, and, arguably, the ability to acquire, maintain and use symbolic knowledge.

Our long term research goal is to improve the performance of robotic systems by addressing the above three limitations, and, specifically to devise "Knowledgeable Robots" [1]. Many researchers are addressing the above challenges. For example, recently, there has been a significant progress in new sensing devices and improved perception capabilities. At the same time, several researchers have addressed forms of collaboration between humans and robots, and voice interaction in particular, that provides a natural approach to communicate with users. Indeed, speech interfaces have reached a level of performance that makes them deployable on hand held devices for applications that go beyond the basic telephone functionality.

Still, few works address the use of classical symbolic methods for knowledge representation and reasoning on robots. Moreover, when this is addressed, the representation of symbolic knowledge is typically based on a static representation of general knowledge (i.e. for action planning), which is disconnected from both the perception system and the interaction with the user.

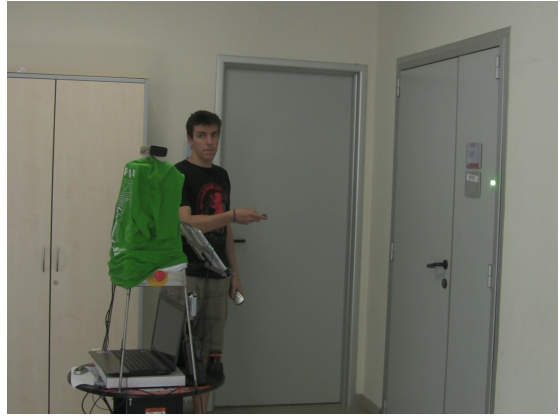

**Fig. 1.** Example of human-robot collaboration: speech-guided tagging of the environment.

We propose to address the challenge of knowledgeable robots by embracing the paradigm of symbiotic autonomy [2], which relies on the interaction with the user in order to overcome the limitations of the robot in perception and action capabilities.

A key issue in the interaction with robots to establish a proper relationship between the symbols used in the representation and the corresponding elements of the operational environment. This process is usually referred to as *symbol grounding* and, in the case of robots, requires to "close the loop" with perception. Consequently, symbol grounding characterizes one of the most critical design issues in the interaction with robots, namely the connection between numeric and symbolic representation.

Specifically, our work addresses the problem of building high-level representations of the environment that embody both metric and symbolic knowledge about it. To this aim, a multi-modal interaction (including a speech interface) can be used in the process of building a very rich representation, by acquiring information in a suitable and natural way.

In order to obtain a representation that the robotic system can effectively use to perform complex tasks, such as "go near the fridge" or "check whether the TV set is on", metric and semantic information need to be suitably merged. The resulting, integrated representation should enable the system to perform topological navigation, understanding target locations and the position of objects in the environment.

The approach that we present in this paper aims at building a rich representation of the robot's knowledge about the environment, comprising a suitable integration of metric and semantic information. More specifically, we describe the representation and the acquisition of the robot's knowledge about a specific environment with the use of a multi-modal human-robot interaction system. We also consider the possibility of extending the approach to an on-line construction and maintenance of the semantic map through a continuous long-standing interaction with the user.

As a general difference with previous work in producing human augmented maps, as well as in representing the knowledge of the robot about the environment, we aim at a rich and detailed representation of a specific environment, as opposed to trying to embed in the robot the knowledge that allows it to operate in any kind of environment, without any previous interaction with the user and the actual operational environment. This is currently possible only through an adequate use of multi-modal natural user interaction (not requiring specific robotic expertise) and, specifically on the use of voice interaction to create a common reference for the symbols representing the knowledge of the robot.

The approach described in this paper builds on previous work for off-line building of a semantic map [3], based on two basic components: 1) a subsystem for Simultaneous Localization And Mapping, that provides a metric map of the environment; 2) a multi-modal interface, including speech, that allows the user to point at the elements of the environment and to assign their semantic role (see Figure 1). The novel contribution here is the definition of a richer representation of the environment, including a Cell Map and a Topological Graph, and of the corresponding procedures for knowledge construction, that increase the capability of the robot to perform complex reasoning for executing complex tasks.

The paper is organized as follows. Section 2 describes the background and a suitable context for our work with respect to state of the art. The representation of the robot's knowledge is presented in Section 3, while the method for its construction is described in Section 4. Conclusions are drawn in the last section.

## 2 Related Work

Hertzberg and Saffiotti [4] characterize semantic knowledge in robotics by two properties: *(i)* the need for an *explicit representation* of knowledge inside the robot; and *(ii)* the need for *grounding the symbols* used in this representation in real physical objects, parameters, and events. Many robotic systems nowadays embody some sort of semantic knowledge, but it is often hard-coded in their implementation. This greatly reduces its reuse, the possibility to reason on it, or to share it with other entities (e.g. robots, hardware devices, or humans). Several works have addressed symbolic representation of knowledge for complex robot tasks. For example, Schiffer et al. [5] describe the implementation of an intelligent service robot whose knowledge about the task to be executed is represented with the logic language READYLOG (a variant of Golog). Aker et al. [6] use Answer Set Programming for representing and reasoning about the actions to be accomplished by a housekeeping robot. De Giacomo et al. [7] present a knowledge representation and reasoning framework based on Description Logics for a service robot. In all these works the knowledge base is assumed to be given to the robot and the methods used to actually acquire this knowledge are not discussed. In this paper we focus on the construction of a semantic representation of knowledge, more precisely on *semantic mapping*.

"A semantic map for a mobile robot is a map that contains, in addition to spatial information about the environment, assignments of mapped features to entities of known classes" [8]. Two properties come out from this definition: 1) semantic maps are independent from the adopted representation for the geometric map behind it: 2D maps, 3D maps, topological maps, texture-based maps; 2) semantic maps also include commonsense knowledge about generic properties of the labeled entities.

The works on semantic mapping can be grouped into two main categories, by distinguishing automatic processing from approaches involving a human user to help the robot in the semantic mapping process.

In the first category, we consider works describing *fully automatic* methods in which human interaction is not considered at all. A first set of techniques aim at extracting features of the environment from laser based metric maps to support labeling and extract high-level information. These works include determining attributes of rooms [9], doorways detection [10], etc. Moreover, in Galindo *et al.* [11], environmental knowledge is represented by augmenting a topological map (extracted with fuzzy morphological operators) with semantic knowledge using anchoring. A second set of techniques make use of classification and clustering for automatic segmentation and labeling of metric maps. For example, in Nücther *et al.* [12], environmental knowledge is extracted by labeling 3D points through the gradient difference between neighboring points; points are then classified into floor-points, object-points, or ceiling-points. While generation of 2D topological maps from metric maps has been described in [13] and [14] (using AdaBoost), in Brunskill et al. [15] (using spectral clustering), in [16] (using Voronoi random fields), etc. Finally, a third set of techniques for object recognition and place categorization use visual features, such as in [17], or a combination of visual and range information, provided from an RGB-D camera, such as in [18]. Although significant progress has been made in fully automated semantic mapping, the approach still suffers from errors and lack of generality.

In the second category, we consider approaches for *human augmented mapping* where the user actively supports the robot to acquire the required knowledge about the environment. In particular, the user role is in grounding symbols to objects that are still autonomously recognized by the robotic platform. In this case, the human-robot interaction is uni-modal, and typically achieved through speech. In Diosi *et al.* [19] an interactive SLAM procedure and a watershed segmentation are employed to create a contextual topological map. In Zender *et al.* [20] a system to create conceptual representations of human-made indoor environments is described. A robotic platform owns a priori knowledge about spatial concepts, and through them builds up an internal representation of the environment acquired through low-level sensors. The user role throughout the acquisition process is to support the robot in place labeling. However, once achieved, the conceptual representation is also useful for effective human-robot dialogue. Pronobis and Jensfelt [21] present a multi-layered semantic mapping algorithm that combines information about the existence of objects and semantic properties about the space, such as room size, share and appearance. These properties decouple low-level information from high-level room classification. The user input, whenever provided, is integrated in the system as additional properties about existing objects. Finally, Nieto-Granda *et al.* [22] adopts human augmented mapping based on a multivariate probabilistic model to associate a spatial region to a semantic label. A user guide supports a robot in this process, by instructing the robot in selecting the labels. Few approaches aim at a more advanced form of human-robot collaboration, where the user actively cooperates with the robot to build a semantic map, not only for place categorization and labeling, but also for object recognition and positioning. Such an interaction is more complex and requires natural paradigms, not to result in a tedious effort for a non-expert user. For this reason, multi-modal interaction is preferred, to naturally deal with different types of information. For example, Kruijff *et al.* [23] introduce a system to improve the mapping process by clarification dialogues between human and robot, using natural language; Randelli *et al.* [3] propose a rich multi-modal interaction, including speech, gesture, and vision enabling for a semantic labeling of environment landmarks that makes the knowledge about the environment actually usable, but without creating a suitable representation.

The approach presented in this paper aims at improving human-robot collaboration by allowing non-expert operators to generate rich semantic information about an environment. To this end, we adopt a multi-modal natural interaction framework to build semantic maps, where the acquired knowledge can effectively support the execution of complex tasks. The main differences with previous work on semantic mapping are: 1) a much richer representation of the environment including semantic descriptions of places, objects and functionalities of the objects; 2) an increased flexibility of the system that can be effectively used in different kinds of environments without requiring a substantial training to build a priori knwowledge.

## 3 Representation of the Robot's Knowledge

In this section we describe the representation of the robot's knowledge, while the description of how this knowledge is acquired and how the representation is actually built is provided in the next section.

The robot's knowledge is divided in two layers: 1) the specific knowledge that the system acquires about the environment, denoted as *world knowledge*; 2) the general knowledge about the domain and its use, that we denote as *domain knowledge*.

It is important to point out that, while the two components may resemble the extensional and intensional components of a classical knowledge base, here they are independent of each other. In fact, the world knowledge may be inconsistent with the domain knowledge (e.g., the robot may have discovered a fridge in the living room, rather than or in addition to the one in the kitchen, while the domain knowledge states that "fridges are in the kitchen"). Generally speaking, domain knowledge (which represents an a priori knowledge about the environment) is used to support the action of the robot, only when specific world knowledge is not available. For example, the robot is able to reach an object (e.g., the fridge in the living room) , because it has been previously discovered; if there is no specific knowledge about the position of an object (e.g., a fridge), the system will refer to the general domain knowledge to find out a possible location (e.g., the kitchen).

More specifically, the robot's knowledge is defined by the following structures:

- World Knowledge
    - Metric Map
    - Instance Signature Data Base
    - Cell Map
    - Topological Graph
- Domain Knowledge
    - Conceptual Knowledge Base

## 3.1 Knowledge about the robot's world

As already mentioned, a semantic map is typically characterized as a low-level representation of the 2D map (i.e. a grid), that is labelled with symbols (e.g. [21], [20], [22]). Such symbols denote for example, the area corresponding to a kitchen (or simply to a room) or the presence of structural elements of the environment and objects. Our representation builds on a similar structure, but it supports a much more detailed description, based on the interaction and hints provided by the user.

In the following definitions we use the term $\mathscr{C}oncept$ to refer to a set of symbols used in the conceptual KB to denote general concepts (i.e., abstraction of objects or locations). For example, *Fridge* and *Kitchen* are concept names of the general Domain Knowledge discussed in the next subsection. While the term $\mathscr{L}abel$ is used to refer to a set of symbols indicating specific instances of objects or locations. For example, *fridge*1 is a label denoting a particular fridge and *kitchen*1 is a label denoting a particular kitchen. In the notation used in this paper, concept names have the first letter capitalized, while labels are denoted with all lowercase letters, typically followed by a digit. The associations between labels and concepts are denoted as $label \mapsto Concept$. For example, $fridge1 \mapsto Fridge$ denotes that the label $fridge1$ is related to the concept *Fridge* (i.e., $fridge1$ is a fridge). Notice that the meaning of the labels is simply that of a pointer to the concept that in the domain knowledge represents the general knowledge about the object or location, and it is not an instance of the concept. In this way, a labeled object can be enriched with general domain information, but it is not required to be consistent with the domain KB.

The representation formalism of the World Knowledge contains the following elements.

**Metric Map.** The *Metric Map* is represented as an occupancy grid generated by a SLAM method. This map has usually a fine discretization (e.g., 5 cm) and is used for low-level robot tasks, such as localization and navigation.

**Instance signatures.** The *instance signatures* are represented as a data base of structured data, where each instance has a unique label ($l \in \mathscr{L}abel$), an associated concept ($C \in \mathscr{C}oncept$) such that $l \mapsto C$, and a set of properties (including for example the position in the environment) expressed as attribute-value pairs.

**Cell Map.** The *Cell Map* is represented as a discretization of the environment in cells of variable size. Each cell represents a portion of a physical area and is an abstraction of locations that are not distinguishable from the point of view of robot high-level behaviors. The Cell Map also includes a function $f : Cell \to 2^{\mathscr{L}abel}$, that maps each cell to a set of labels associated to concepts in the conceptual KB of the Domain Knowledge, and a connectivity relation $Connect \subseteq Cell \times Cell$, that describes the connectivity between adjacent cells.

**Topological Graph.** The *Topological Graph* is a graph where nodes are locations associated to cells in the Cell Map and edges are connections between these locations. Locations are distinguished in two types: static and dynamic. For the static locations, the corresponding positions (i.e., the correspondences with the metric map) are fixed, while in the dynamic locations, the corresponding positions are variable within a given area of the environment. Since the Topological Graph is used by the robot for navigation purposes, the edges also contains the specific navigation behavior that is required for the robot to move from one location to another. In this way the topological map is also used to generate appropriate sequences of behaviors to achieve the robot's navigation goals.

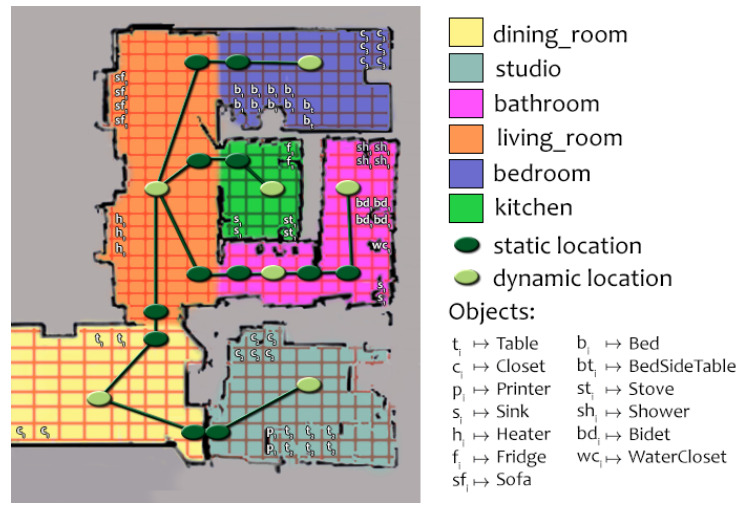

**Fig. 2.** An example of World Knowledge.

In order to better explain these representations, in the remaining of this section we provide an example of World Knowledge for a home environment, while in the next section we describe how this representation has been built and how the knowledge has been acquired.

Figure 2 shows a graphical representation of the World Knowledge as described above. The Metric Map with a resolution of 5 cm generated by the SLAM method is shown as background of the image, where black pixels represent occupied cells.

Every object and location reported in the figure has an entry in the data base of instance signatures. In particular, each instance is determined by a unique label representing an entity within the mapped environment, (e.g. $fridge1$), the corresponding concept according to the conceptual KB (e.g. $f_1 \mapsto Fridge$), and a set of properties (e.g., position = $<x, y, \theta>$, color = *white*, open = *false*, ...).

The Cell Map is shown in the same image. Cells are delimited by solid borders; colors and text in the cell specify the labels associated to each cell. The labels in the Cell Map refer to a typical area of a home (kitchen, living room, bathroom, bedroom, etc.) and to typical objects (tables, chairs, heaters, doors, etc.). Each cell can have more than one label, thus representing that the same area can be associated with more concepts. In particular, each color corresponds to a label indicated in the legend to the right and text (where present) corresponds to additional labels for that cell corresponding to the objects reported on the right side of the figure. For example, the green cells labeled with $f_1$ (top-right corner of the kitchen area) are mapped to the labels $\{f_1, kitchen\}$ and these labels are associated with the concepts *Fridge* and *Kitchen*, i.e. $f_1 \mapsto Fridge$ and $kitchen \mapsto Kitchen$, respectively. Thus, the corresponding area is characterized as being occupied by a fridge and as belonging to the kitchen. Connectivity relations between cells are not explicitly shown in the figure, but they can be derived by adjacent cells.

The Topological Graph, that establishes the relationship between the metric and the symbolic representation, is shown in the figure as a graph connecting oval nodes. Dark nodes are static locations, while light nodes are dynamic locations, that are associated to the main areas (labels) of the environment. The static locations denote specific positions that are of interest for the robot tasks (e.g., the position to enter in a room), while the dynamic locations are used to denote areas, where instantiation of the position (i.e., the mapping to the metric map) for a navigation behavior is executed at run-time, depending on the current status of the robot and on its goals.

### 3.2 General knowledge about the domain

In previous work, domain knowledge has typically been characterized as a conceptual knowledge base, representing a hierarchy of concepts, including their properties and relations, a priori asserted as representative of any environment. The conceptual knowledge base contains a taxonomy of the concepts involved in the environment tied by an *is-a* relation, as well as their properties and relations (see [11], [13]). These concepts are used in the World Knowledge to characterize the specific instances of the environment, as explained in the previous section. In a typical representation three top-most classes have been considered: *Areas*, *Structural Elements*, and *Objects*. *Areas* denote places in the environment (corridors, rooms, etc.), *Structural Elements* are static entities that form the environment and that topologically connect areas (windows, doors, etc.), while *Objects* are elements in the environment not related to its structure and located within areas (printers, tables, etc.). A snapshot of the conceptual knowledge base used in the experiments is reported in Figure 3.

The knowledge base contains also environmental properties of the defined concepts, like size and functionalities of objects, typical connections among places, etc., that are useful to describe the general knowledge about an environment and to support robot task execution. For example, the general knowledge in a home environment that fridges are in kitchens helps the robot to perform tasks related with a fridge even when the exact location of a fridge is not known. A noticeable difference with respect to previous works is that the purpose of this component is not to classify spaces and objects; rather, we focus on the inheritance of properties to support the map acquisition, as well as different actions of the robot in the environment.

The uses of the Domain Knowledge base are manifold. Spatial properties of objects can be used to build metric representations of objects (see next section); functional properties can be used to determine location of object (e.g., for devising a plan to search for an object); physical properties can be used to check preconditions for executing transportation, etc.

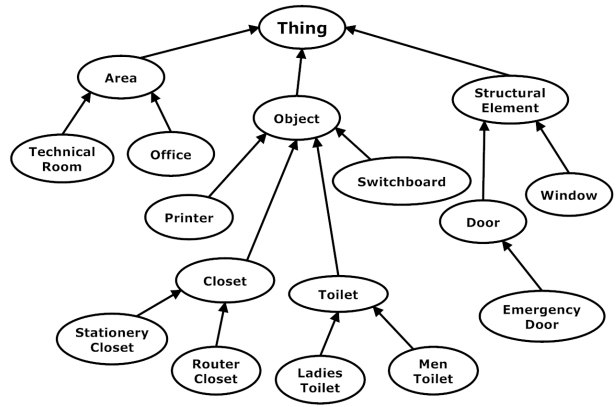

**Fig. 3.** A fragment of the conceptual knowledge base used in the experiments.

As a matter of fact, any source of domain knowledge may be accessed and used by the robot, such as for example in querying the web, (e.g. RoboEarth [24] and OpenEval[25]), or even querying other robots operating in similar environments.

## 4 Acquisition of the Robot's Knowledge

The process of building the representation of the robot's knowledge is composed of two phases that take place, with the help of the user, after the exploration of the environment: 1) *Metric Map and Instance Signatures Construction*. A 2D metric map is generated through a SLAM module and the set of instance signatures is created; 2) *Cell Map and Topological Graph Generation*. Starting from the 2D metric map, a grid-based topological representation (*cell map*) is obtained. Then, the topological graph needed by the robot to perform high level behaviours is built by processing the instance signatures and the cell map.

### 4.1 Metric Map and Instance Signatures

In the first phase of the knowledge acquisition procedure (see Fig. 4), the robot is used to navigate the environment in order to create the 2D map and to register the positions of the different objects of interest. The user can tag a specific object by using a commercial laser pointer (Fig. 5a).

While the object is pointed through the laser, the user has to name the object, so that the label can be assigned to it. The image segmentation module is responsible for extracting the 3D shape of the pointed object by exploiting the RGB-D data coming from the Kinect, sensor [26]. The laser dot is detected in the RGB image (Fig. 5b) by searching for the color of the light emitted (green in our case). Then, all the planes in the scene are extracted from the 3D point cloud and those that do not contain the dot are discarded. The remaining points are analysed to segment the shape of the object of interest (Fig. 5c).

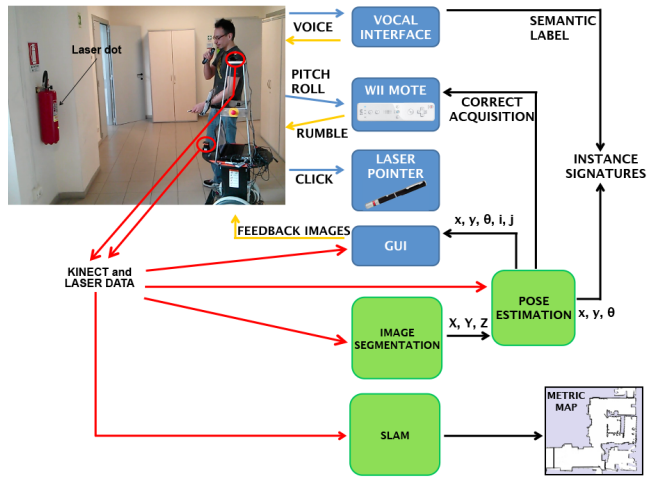

**Fig. 4.** Input data are obtained from a human user and a robot in order to produce the metric map and the instance signatures.

It is worth noticing that the image segmentation module exploits both the information provided by the data base (height, width, and length) and the laser dot position in the Kinect view (used as seed point of the expansion). If the extracted shape is coherent with the dimensional attributes of the corresponding instance signature stored in the data base (i.e., the one with the same label recognized by the vocal interface), then a correct acquisition message is sent to the user in order to acknowledge the successful association.

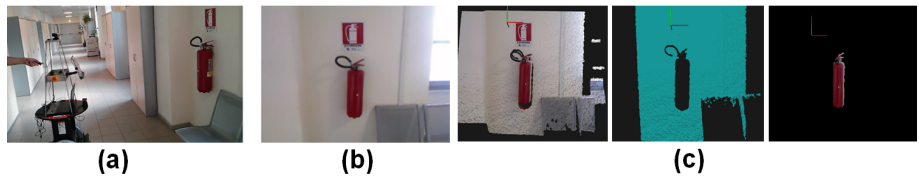

(a) (b) (c)

**Fig. 5.** Object tagging. a) The object is tagged with the laser pointer. b) The RGB image from the Kinect is used to locate the laser dot. c) The 3D point cloud (left) is analysed to find all the planes, the ones do not containing the laser dot are discarded (center) and the dot is used as seed point to segment the tagged object (right).

The pose estimation module outputs the 2D position $x, y$ and the bearing $\theta$ of the tagged object. The pose $(x, y, \theta)$ is calculated by taking into account the normal corre-sponding to the surface of the segmented object in reference frame of the Kinect and then applying a transformation to the reference frame of the robot.

When the acquisition run is terminated all the recorded data (i.e., laser scans and object poses) are processed. The laser scans are used as the input for generating the 2D metric map. A Graph-based SLAM approach, as presented in [27], is used to estimate the 2D map of the environment and the trajectory of the robot. The nodes in the graph represent the poses of the robot, while the edges express spatial constraints arising from odometry measurements. In order to obtain the object map that integrates the instance signatures into the SLAM generated map, new edges are added to the graph together with the previous ones (given by laser scans and robot odometry). Each registered object pose is considered as a new edge of the graph, embodying the pose and the label of the object. When the new expanded graph is complete, a least-square minimization algorithm [28] is performed to obtain the optimized final 2D metric map. The final poses of the tagged objects (coherent with the optimized map) are stored as instance signatures together with corresponding labels.

### 4.2  Cell Map and Topological Graph

In the second phase of the knowledge acquisition procedure the cell map and the topological graph are generated (see Fig. 6).

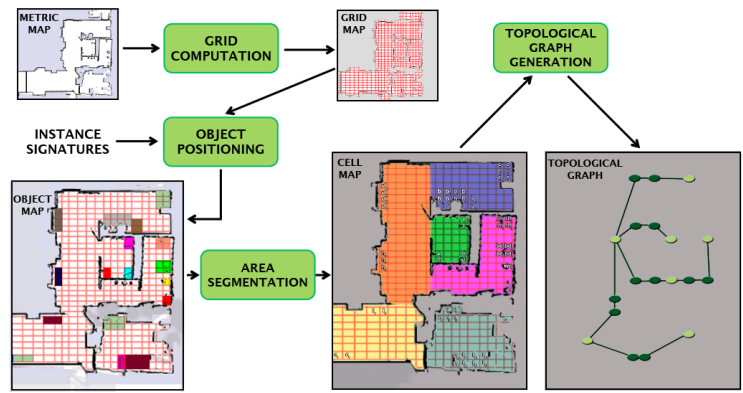

**Fig. 6.** Cell map and topological graph generation. The metric map is enriched with semantic information to obtain the cell map. The instance signatures and the cell map are used to create a topological graph.

The cell map contains a high-level description about the regions, structural elements, and objects contained in the environment. It is generated using both the instance signatures and the 2D topological map. The algorithm to generate such a map can be further decomposed into the following steps: *1)* rasterizing the metric representation of the map into a grid-based topological representation (grid computation); *2)* including in this new map representations of the objects and structural elements tagged in the environment (object addition); *3)* automatically labeling the areas of the environment (labeling areas), using contour closure and region filling techniques.

*Grid computation* is accomplished by applying the Hough Transform at different resolution levels to identify the lines passing through the walls. In such a way, it is possible to find all the main walls in the map. The cells of the grid have different sizes with respect to the amount of relevant features in the considered area (for example, the cells in the corridor are wider then the ones in the offices).

*Object addition* consists in placing the tagged objects into the grid map. For this we use objects' pose, shape and size, present in the instance signature data base.

Finally, *labeling areas* is applied to segment the environment into different rooms. The algorithm is enhanced by considering the tagged objects and structural elements previously included in the environment. For example, a simple heuristic uses the knowledge about doors to separate different areas.

In the final step of our knowledge building process, a topological graph is created by using the knowledge contained both in the cell map and in the instance signatures. Such graph embodies the information needed to the robot for navigating and acting in the environment. The constituting nodes of this graph are locations associated to cells in the cell map, while the edges are connections between these locations. For each room, two kinds of nodes are produced: static and dynamic. Static nodes represent locations that have a fixed position on the map and they are used to deal with critical pathways in the map (i.e. doors), where a specific behavior is required. Dynamic nodes represent, instead, variable positions within a given area of the environment. Thus, for each room mapped in the cell map, a dynamic node is created while, for each doorway that connects such a room to another, a static node is created with its fixed position set in front of the considered door. Finally, each static node is connected to the dynamic node of the room it belongs to and to the static node that represents the location in front of the other side of its related door. The construction of the topological graph highlights a twofold aspect of our system: both the knowledge provided through user interaction and a-priori high-level knowledge contribute to refine the overall mapping process.

## 5   Conclusions

In this paper, we propose an approach to build an effective representation of the knowledge about the environment where the robot operates, by relying on multi-modal human-robot interaction, including natural language. The proposed framework makes it possible to bridge the gap between the numerical and the symbolic knowledge needed by the robot, by grounding the symbols onto the objects represented in the metric map. As a result, the proposed setting for representing the knowledge of the robot is significantly richer than previous approaches in the literature; in particular, it supports better performance and greater generality with respect to fully automatic methods. Through the acquired representation the robot can execute complex tasks. For example, the robot can execute complex commands such as "check whether in the corridor the third window on the left is open", retrieve its metric position in the map, plan a path and reach the specified location and execute the task.

The proposed approach can be further developed in at least two fundamental directions, that we are currently investigating. First, the generation of the map is currently done off-line. A more general setting should enable the robot to grow and update the representation through a continuous interaction with the user. While the proposed approach to symbol grounding can be naturally extended, the new framework would require to ensure the consistency of the representation and to deal with the changes in the environment and the dynamics of objects. Second, currently the robot can only learn knowledge about the environment, but not knowledge about tasks to be accomplished in response to user commands. This extension would require a representation of the robot's actions and an extended dialogue to acquire input from the user.

In conclusion, in our view, the proposed approach brings about a new perspective and new developments for the representation of the robot's knowledge.

## References

[1] Aiello, L.C., Nardi, D., Randelli, G., Scalzo, C.: Suppose you have a robot. Gerhard Lakemeyer and Sheila A. McIlraith (Eds.) Knowing, Reasoning, and Acting, Essays in Honour of Hector J. Levesque (2011) 27–46
[2] Rosenthal, S., Biswas, J., Veloso, M.: An effective personal mobile robot agent through symbiotic human-robot interaction. In: Proc. of 9th International Joint Conference on Autonomous Agents and Multi-Agent Systems (AAMAS). (2010)
[3] Randelli, G., Bonanni, T.M., Iocchi, L., Nardi, D.: Knowledge acquisition through human-robot multimodal interaction. Intelligent Service Robotics **6** (2013) 19–31
[4] Hertzberg, J., Saffiotti, A.: Using semantic knowledge in robotics. Robotics and Autonomous Systems **56**(11) (2008) 875–877 Semantic Knowledge in Robotics.
[5] Schiffer, S., Ferrein, A., Lakemeyer, G.: Caesar: An intelligent domestic service robot. Intelligent Service Robotics **5(4)** (2012) 259–273
[6] Aker, E., Patoglu, V., Erdem, E.: Answer set programming for reasoning with semantic knowledge in collaborative housekeeping robotics. In: Proc. of IFAC SYROCO. (2012)
[7] De Giacomo, G., Iocchi, L., Nardi, D., Rosati, R.: A theory and implementation of cognitive mobile robots. Journal of Logic and Computation **5**(9) (1999) 759–785
[8] Nüchter, A., Hertzberg, J.: Towards semantic maps for mobile robots. Robot. Auton. Syst. **56**(11) (November 2008) 915–926
[9] Buschka, P., Saffiotti, A.: A virtual sensor for room detection. In: Proceedings of the IEEE/RSJ International Conference on Intelligent Robots and Systems (IROS). (2002) 637–642
[10] Anguelov, D., Koller, D., Parker, E., Thrun, S.: Detecting and modeling doors with mobile robots. In: Proc. of the IEEE International Conference on Robotics and Automation (ICRA). (2004)
[11] Galindo, C., Saffiotti, A., Coradeschi, S., Buschka, P., Fernández-Madrigal, J., González, J.: Multi-hierarchical semantic maps for mobile robotics. In: Proceedings of the IEEE/RSJ International Conference on Intelligent Robots and Systems (IROS), Edmonton, CA (2005) 3492–3497 Online at http://www.aass.oru.se/˜asaffio/.
[12] Nüchter, A., Wulf, O., Lingemann, K., Hertzberg, J., Wagner, B., Surmann, H.: 3D Mapping with Semantic Knowledge. In: RoboCup 2005: Robot Soccer World Cup IX. (2005)
[13] Martínez Mozos, O., Burgard, W.: Supervised learning of topological maps using semantic information extracted from range data. In: Proceedings of the IEEE/RSJ International Conference on Intelligent Robots and Systems (IROS), Beijing, China (2006) 2772–2777

[14] Goerke, N., Braun, S.: Building semantic annotated maps by mobile robots. In: Proceedings of the Conference Towards Autonomous Robotic Systems, Londonderry, UK. (2009)

[15] Brunskill, E., Kollar, T., Roy, N.: Topological mapping using spectral clustering and classification. In: Proc. of IEEE/RSJ Conference on Robots and Systems (IROS). (2007)

[16] Friedman, S., Pasula, H., Fox, D.: Voronoi random fields: Extracting the topological structure of indoor environments via place labeling. In: Proc. of 19th International Joint Conference on Artificial Intelligence (IJCAI). (2007)

[17] Wu, J., Christenseny, H.I., Rehg, J.M.: Visual place categorization: Problem, dataset, and algorithm. In: Proc. of IEEE/RSJ Conference on Robots and Systems (IROS). (2009)

[18] Mozos, O.M., Mizutani, H., Kurazume, R., Hasegawa, T.: Categorization of indoor places using the kinect sensor. Sensors **12**(5) (2012) 6695–6711

[19] Diosi, A., Taylor, G., Kleeman, L.: Interactive slam using laser and advanced sonar. In: Proceedings of the IEEE International Conference on Robotics and Automation, Barcelona, Spain (2005) 1103–1108

[20] Zender, H., Martínez Mozos, O., Jensfelt, P., Kruijff, G., Burgard, W.: Conceptual spatial representations for indoor mobile robots. Robotics and Autonomous Systems **56**(6) (2008) 493–502

[21] Pronobis, A., Jensfelt, P.: Large-scale semantic mapping and reasoning with heterogeneous modalities. In: Proceedings of the 2012 IEEE International Conference on Robotics and Automation (ICRA'12), Saint Paul, MN, USA (May 2012)

[22] Nieto-Granda, C., III, J.G.R., Trevor, A.J.B., Christensen, H.I.: Semantic map partitioning in indoor environments using regional analysis. In: 2010 IEEE/RSJ International Conference on Intelligent Robots and Systems, October 18-22, 2010, Taipei, Taiwan, IEEE (2010) 1451–1456

[23] Kruijff, G., Zender, H., Jensfelt, P., Christensen, H.: Clarification dialogues in human-augmented mapping. In: Proceedings of the 1st Annual Conference on Human-Robot Interaction (HRI'06), Salt Lake City, UT (March 2006)

[24] Waibel, M., Beetz, M., Civera, J., D'Andrea, R., Elfring, J., Galvez-Lopez, D., Haussermann, K., Janssen, R., Montiel, J., Perzylo, A., Schiessle, B., Tenorth, M., Zweigle, O., van de Molengraft, M.: RoboEarth, - a world wide web for robots. Robotics and Automation Magazine **18**(2) (2011) 69–82

[25] Samadi, M., Kollar, T., Veloso, M.: Using the web to interactively learn to find objects. In: Proc. 26th Conference on Artificial Intelligence (AAAI). (2012)

[26] Bonanni, T.M., Pennisi, A., Bloisi, D.D., Iocchi, L., Nardi, D.: Human-robot collaboration for semantic labeling of the environment. In: Proc. of the 3rd Workshop on Semantic Perception, Mapping and Exploration (SPME). (2013) 1–6

[27] Grisetti, G., Kuemmerle, R., Stachniss, C., Burgard, W.: A tutorial on graph-based SLAM. Intelligent Transportation Systems Magazine, IEEE **2**(4) (2010) 31–43

[28] Kuemmerle, R., Grisetti, G., Strasdat, H., Konolige, K., Burgard, W.: g2o: A general framework for graph optimization. In: Proc. of the IEEE Int. Conf. on Robotics and Automation (ICRA), Shanghai, China (May 2011)